# Prediction of Human Trajectory Following a Haptic Robotic Guide Using Recurrent Neural Networks

Hee-Seung Moon and Jiwon Seo, *Member, IEEE*

*Abstract*—Social intelligence is an important requirement for enabling robots to collaborate with people. In particular, human path prediction is an essential capability for robots in that it prevents potential collision with a human and allows the robot to safely make larger movements. In this paper, we present a method for predicting the trajectory of a human who follows a haptic robotic guide without using sight, which is valuable for assistive robots that aid the visually impaired. We apply a deep learning method based on recurrent neural networks using multimodal data: (1) human trajectory, (2) movement of the robotic guide, (3) haptic input data measured from the physical interaction between the human and the robot, (4) human depth data. We collected actual human trajectory and multimodal response data through indoor experiments. Our model outperformed the baseline result while using only the robot data with the observed human trajectory, and it shows even better results when using additional haptic and depth data.

## I. INTRODUCTION

Although autonomous robots are already performing an increasing number of tasks in human society, it is necessary to improve their social intelligence to entrust them with more significant work. Social intelligence is required for collaborating and engaging with social agents, such as humans and other robots, by understanding their behavioral patterns and limitations [1], [2]. For some social robots, assistive robots, and robot workers, social intelligence has an important influence on their task efficiency because these robots use physical interaction with humans to provide assistance or receive users' responses [3]–[6].

One typical type of social intelligence for these robots is human trajectory prediction, having the goal of generating the future trajectory of people based on their previous trajectory. Specifically, when a robot cooperates with a human, accurate human trajectory prediction can achieve following advantages: (1) robots can prevent the user from getting into accidents using the robots' sensory information, (2) by knowing the user's behavior space, the robot can play a more active role, and (3) robots can make an appropriate motion that induces an intuitive human reaction that does not put stress on a person's muscles.

However, predicting a human path is a challenging problem in general because it requires an understanding of human dynamics and human reaction within social interactions between human-robot or human-human. Further developing an early method [7] that used manually derived functions to define the relationship between a person's social interactions and the path toward their goal, researchers recently made significant progress by applying the deep learning method for human trajectory prediction in crowded spaces [8]–[10]. To solve such a challenging problem, researchers designed a learning architecture based on recurrent neural networks (RNNs), which have the ability to learn and generate sequential data using the pedestrian's observed trajectory, the location of nearby people [8], [9], and the surrounding terrain information as training data [9], [10].

In this paper, we address the problem of predicting human trajectory specifically for the situation where the human follows a haptic robotic guide without using sight. This type of human-robot team [11]–[14] is highly valuable because it allows visually impaired people or people in a visually restricted situation, such as a disaster area, to navigate through unknown spaces. Within this human-robot formation, the prediction problem will be quite different from previous cases where the human moves freely in open space; this is because the human is now instructed by the physical assistant to take the next step. For example, a person without sight internally anticipates where the robot is leading them in order to more naturally make their next move; therefore, the movement of the person is affected by both the actual physical assistance and the person's internal anticipation. Considering this, it is still a challenging problem to predict an accurate trajectory of the human following the robot.

To predict an accurate human path, we apply the deep learning method based on four types of multimodal sequential data: (1) human trajectory, (2) movement of the robotic guide, (3) haptic input data measured from the physical interaction between the human and the robot, and (4) human depth data.

This research was supported by the Basic Science Research Program through the National Research Foundation of Korea (NRF) funded by the Ministry of Education (NRF-2018R1D1A1B07043580). This research was also supported by the Ministry of Science and ICT (MSIT), Korea, under the "ICT Consilience Creative Program" (IITP-2018-2017-0-01015) supervised by the Institute for Information & Communications Technology Promotion (IITP).
H. S. Moon and J. Seo are with the School of Integrated Technology and the Yonsei Institute of Convergence Technology, Yonsei University, Incheon 21983, Republic of Korea (e-mail: {hs.moon, jiwon.seo}@yonsei.ac.kr).

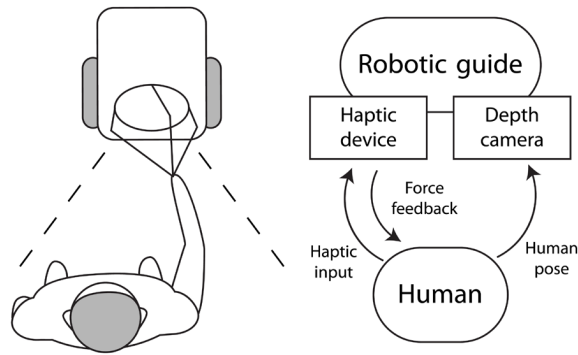

Figure 1. Overview of our robotic-guide system. The human follows the haptic robotic guide without using vision. The dotted line indicates the angle of view of the depth camera.

To measure accurate haptic input from the user, our robotic-guide system is equipped with a haptic device, and the user is instructed to follow the robotic guide while holding the haptic device (Fig. 1). In addition, a depth camera captures the human's pose, and latent vectors are extracted from the depth image using a pre-trained variational autoencoder (VAE). Finally, the multimodal data are used as input to our RNN model so that the future trajectory of the human can be generated from the model.

## II. RELATED WORK

### A. Haptic Robotic Guide

Haptic sensation can effectively convey information to users whose vision is blocked or overloaded [15], [16]; therefore, it can also be used as the primary means of the human-robot interaction [3]–[6], [17]. In particular, because haptic feedback can provide directional cues, it has been used in several applications where a robot gives directional guidance to a visually impaired pedestrian. There are two types of haptic guidance: (1) tactile feedback, which is felt through the skin, such as vibration or texture, and (2) kinesthetic feedback, which is felt through joints and muscles. The guidance using tactile feedback is typically conducted by connecting multiple vibrotactile motors to the user through a belt [13] or armbands [14] and conveying directional information by operating the motor at the corresponding position. Providing kinesthetic feedback allows for more aggressive guidance because it can directly apply force to the user. In [11], a room-sized haptic interface was used, and the user received either active guidance, which pulled the user toward the target path, or passive guidance, which prevented the user from leaving the target path. In [12], through a spring system connecting the robotic guide and a hand-held stick, the user was guided by the spring tension according to the angle of the stick.

### B. Human Motion Prediction Using RNNs

The recurrent neural network (RNN) and its improved models, long short-term memory (LSTM) [18] and gated recurrent unit (GRU) [19], are designed to process time sequential data as inputs or outputs. Recent research shows that those recurrent models are effective for predicting future human motion [20]–[22]. Specifically in [22], the authors showed improved performance in predicting the next movements of people using single-layered GRUs with residual connections compared to that using multi-layered LSTM architectures from [21].

In terms of human path prediction, the problem of how pedestrians move in an open crowded space has been mostly studied with the LSTM-based method. The researchers attempt to predict the next path of the pedestrian using additional context data rather than simply using the pedestrian's observed path. Social LSTM [8] combines motion information of other people around the pedestrian with their relative positions and feeds them together as input data. SS-LSTM [9] and Scene-LSTM [10] have shown improved results by using scene features extracted from video frames. Meanwhile, to the best of our knowledge, there has been no attempt to predict how a person following the haptic robotic guide would move in the future.

## III. PROPOSED METHOD

### A. Our Haptic Robotic Guide

To acquire multimodal user data for the human path prediction, the haptic robotic guide that was developed in our previous work [23], [24] was used. The robotic guide consists of a Stella B2 mobile robot (NTREX Inc., Korea), an Omega 7 haptic device (Force Dimension, Switzerland), and an Xtion 2 RGB-D camera (ASUS, Taiwan). In this work, the user follows the robotic guide by holding the end effector of the haptic device, and the robot guides them using an elastic system that provides haptic force feedback when the end effector moves away from the origin. Because the movement of the user is limited to a 2D horizontal plane, the end effector is programmed to only move in a 2D horizontal plane and not in the vertical direction. The position of the end effector can be measured with a resolution of less than 0.01 mm, and the haptic feedback is generated using the immediate location of the end effector based on (1), where $x_{ee}$ represents the location of the end effector and $v_{ee}$ represents the velocity of the end effector. We set the spring constant to $k = 500$ N/m and the damping constant to $b = 30$ N·s/m so that

$$F = -kx_{ee} - bv_{ee}. \quad (1)$$

The depth camera, which is mounted on the robotic guide facing backward, acquires a 640 × 480 depth image of the following person's torso. Using a synchronized signal from a robot-mounted laptop computer, haptic input data and depth images are acquired simultaneously four times a second, i.e., 4 frames per second (fps).

### B. Our Model

*Problem Definition:* Our study focuses on the human response according to the interaction between the human and the robot. Therefore, we assume that the position and the orientation of the robotic guide at every time-step is known. The path of the human is tracked using relative position and orientation with the robot ($x^t_{H/R}$, $y^t_{H/R}$, $\theta^t_{H/R}$) (Fig. 2(a)). Thus, our problem can be formulated as observing the human positions ($x^t_{H/R}$, $y^t_{H/R}$, $\theta^t_{H/R}$) at $t = 1$ to $T_{obs}$ and predicting the next human positions at $t = T_{obs} + 1$ to $T_{obs} + T_{pred}$. We also use three additional sources of data for the prediction: (1) robot data, which includes both the velocity command provided to the robot ($v^t_l$, $v^t_r$) and robot movement ($\Delta x^t_R$, $\Delta y^t_R$, $\Delta\theta^t_R$), which is measured as the relative position and orientation of the robot at time $t$ based on the robot at time $t - 1$ (Fig. 2(b)), (2) haptic data, measured as the 2D location of the end effector ($r^t_x$, $r^t_y$), and (3) depth data, which are latent vectors $z^t$ extracted from depth images using pre-trained VAE. It should be noted that we use haptic and depth data, which contain the user's response to physical interaction with the robot, only at $t = 1$ to $T_{obs}$, whereas robot data for all time-steps, $t = 1$ to $T_{obs} + T_{pred}$, is used; this is because we assumed (above) that all of the robot path data is known.

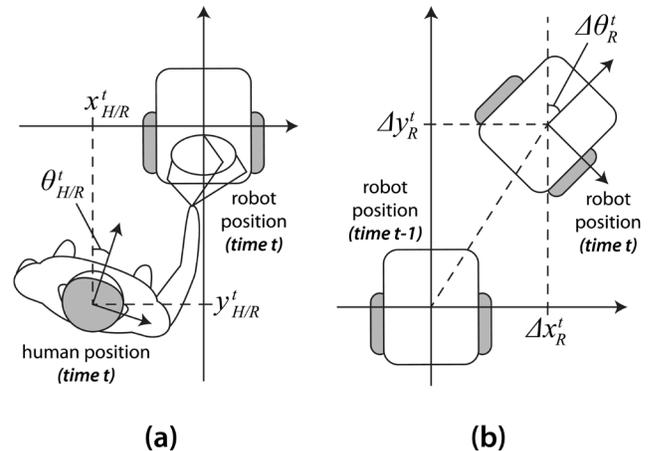

Figure 2. (a) Schematic representation of the relative position and orientation of the human to the robot at time $t$. (b) Schematic representation of the robot movement at time $t$.

*Network Design:* Our recurrent network architecture is composed of double-layered GRUs (Fig. 3), and a sequence-to-sequence structure [25] is applied, where the encoder and the decoder share the same weight as in [22]. At the encoding stage, the observed data sequence ($t = 1$ to $T_{obs}$) is fed into the network, and the model updates its own hidden state vectors and predicts the next time-step data ($t = T_{obs} + 1$). At the decoding stage, our model predicts the next data set one-by-one in each time-step while the last output data is used as the next input. Because the robot data are assumed to be known, our recurrent networks are designed to produce all of the data except for the robot data. Also, by adding the residual connection from input to just before output, the GRUs model the change value for next time-step data.

The operation sequence of the network is as follows. The encoding stage of the pre-trained VAE, which was designed in our previous work [24], extracts latent feature vectors of size 5 from depth images (we will not cover the VAE stage in this paper.) Then, all of the input data at time $t$, which consist of robot data $R^t = (v^t_l, v^t_r, \Delta x^t_R, \Delta y^t_R, \Delta\theta^t_R)$, human position data $P^t = (x^t_{H/R}, y^t_{H/R}, \theta^t_{H/R})$, haptic data $H^t = (r^t_x, r^t_y)$, and depth data $D^t = z^t$ are normalized to values between $-1$ and $1$ and concatenated as follows:

$$Input^t = R^t \oplus P^t \oplus H^t \oplus D^t, \quad (2)$$

where $\oplus$ is the concatenation operator. By passing through GRUs and the linear layer, intermediate vector $C^t$, which represents the change value, is computed as follows:

$$h^t_1 = GRU_1(h^{t-1}_1, Input^t; W_1), \quad (3)$$
$$h^t_2 = GRU_2(h^{t-1}_2, h^t_1; W_2), \quad (4)$$
$$C^t = W_o h^t_2 + b_o, \quad (5)$$

where $W_1$, $W_2$, and $W_o$ represent the weight of $GRU_1$ cell, $GRU_2$ cell, and the linear layer, respectively, and $b_o$ represents the bias of the linear layer. Finally, by passing the residual layer, predicted human position, haptic, and depth data at time $t + 1$ are computed as follows:

$$Output^t = C^t + slice(Input^t), \quad (6)$$
$$(P^{t+1}_{pred}, H^{t+1}_{pred}, D^{t+1}_{pred}) = Output^t, \quad (7)$$

where $slice(\cdot)$ represents the slice operator that removes the robot data part.

*Implementation details:* Before we trained our recurrent model, the learning procedure of the VAE model for depth image processing, which has the same structure as [24] except for the latent vector size, was performed first. We used only the depth images of the training data belonging to the recurrent model for VAE learning. The VAE model was trained for 100 epochs with an Adam optimizer, and the learning rate was set to 0.001. In the recurrent model, we used hidden states of size 64 for both GRU cells. Because the goal of our model is to predict human trajectory, the training of the model was implemented by minimizing the MSE loss function of $P^{t+1}_{pred} = (x^{t+1}_{H/R}, y^{t+1}_{H/R}, \theta^{t+1}_{H/R})_{pred}$. The Adam optimizer was used with the learning rate = 0.001 for 500 epochs. Also, we clipped the parameter gradients to a maximum norm of 5 for stabilized training. Both the VAE and recurrent models were built using Keras, and all of the training procedures were implemented on an NVIDIA GTX 1080Ti GPU.

IV. EXPERIMENTAL SETUP

A. Data Acquisition

We collected the robot-human trajectory and multimodal human response data for model training/validation from an indoor experiment. Six healthy participants (5 males, age range 23–26, all right-handed) were involved in the experiment. In the 6 m × 6 m indoor laboratory environment, the robotic guide was set to move along a random path for 20–30 s in each session, and the participants were instructed to follow a randomly moving haptic robotic guide while blindfolded (Fig. 4). Our study was performed in accordance with the ethical standards laid down by the 1964 Declaration of Helsinki. All participants provided informed written consent. Following the relevant Act and Enforcement Rules, which are specified below, from the Korean Ministry of Health and Welfare, our experimental procedure is exempt from local ethics committee approval. According to Article 15 (2) of the Bioethics and Safety Act and Article 13 of the Enforcement Rule of Bioethics and Safety Act, a research project "which utilizes a measurement equipment with simple physical contact that does not cause any physical change in the subject" is exempt from such approval. Our experimental procedure was designed to contact only a haptic device that does not cause any physical change in the participant.

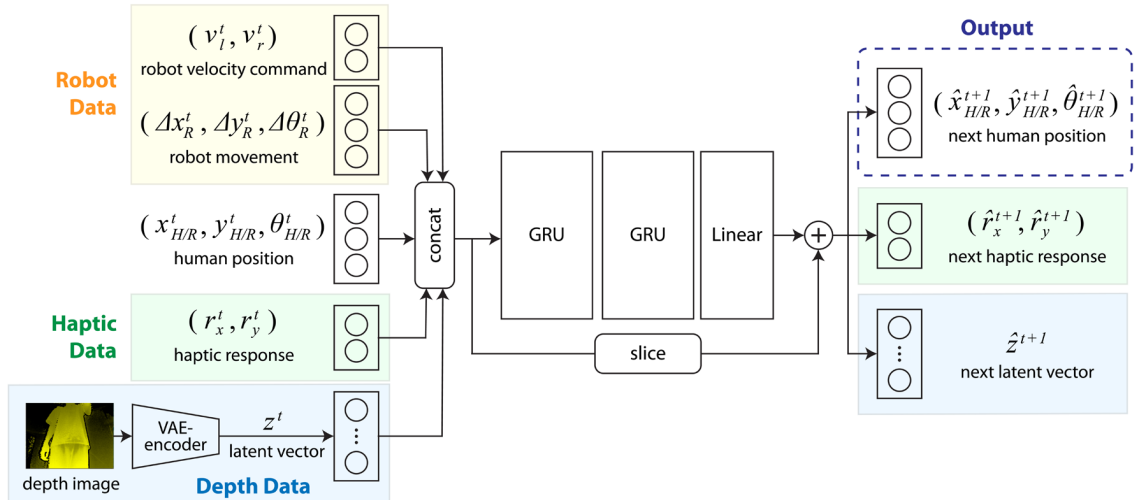

Figure 3. Overview of our model, which consists of double-layered GRUs, one linear layer, and a residual connection. All of the hatted (^) characters indicate that the values are the predicted values. The number of the circles in the box next to the description represents the size of the data.

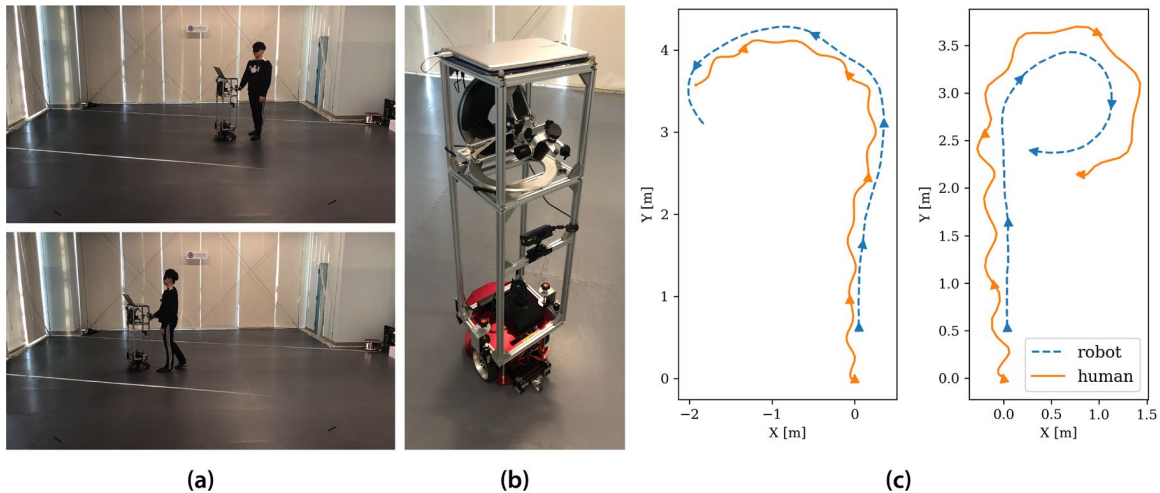

Figure 4. (a) A blindfolded participant is instructed to follow our haptic robotic guide, which moves along a random path. (b) Our haptic robotic guide, which consists of a mobile robot, a haptic device, a depth camera, and motion capture markers. (c) Example trajectories of the randomly moving robotic guide and the following participant.

We obtained actual position and orientation data of the robot and participants by attaching a Vicon motion capture system to each participant and the robot. A 4-hour experiment, which included approximately 150 sessions, was implemented for each participant. In this study, we observed 8 time-step data ($T_{obs}$ = 8) and predicted the next 12 time-step data ($T_{pred}$ = 12), which was a similar process to that of the conventional method in [8]–[10]. Because 1 time-step represented 250 ms in our data acquisition process, our model predicted the next 3-second trajectory by observing 2 seconds of data. In total, we obtained 55,599 data samples by windowing all of the collected data with a length of 20 time-steps.

### B. Evaluation

Because the final result of our model includes predicted human position and orientation data, the metrics of our evaluation is divided into displacement error for position prediction and angle error for orientation prediction. Also, the following two types of the metrics are used for each error.

- *Mean Displacement/Angle Error (MDE/MAE)*: Average distance or angle error between the predicted values and ground truth values at all time-steps.

- *Final Displacement/Angle Error (FDE/FAE)*: Distance or angle error between the predicted values and ground truth values at the final time-step. Therefore, FDE/FAE measures whether the prediction accuracy can be maintained as the predicted time increases compared to MDE/MAE.

For comparison of the performance of the human path prediction, the results using the linear regression model have been selected as a baseline [8]–[10]. However, we cannot apply the linear method for the path prediction following the guidance of the robot, because this method does not take into account the preset path of the robotic guide. To our knowledge, there has been no baseline method used for this particular situation, therefore a newly designed baseline method is implemented. We also compare the results of each multimodal data usage combination as follows. Note that the observed path data of the human is always used, so this is not specified in each method title below.

- *Baseline*: A method designed using the tendency that the human path following the robotic guide is reflected by the path of the robot, albeit delayed (as seen in Fig. 4(c)). The position of the human is estimated assuming that the person moves with the same relative position from the position where the robot was *n* time-steps before, and *n* is determined as the value that minimizes the positional difference between the human and the robot *n* time-steps before in the training data.

- *Using Robot Data Only (R)*: A method using our recurrent model with only robot data in addition to the human trajectory observation.

- *Using Robot and Haptic Data (R+H)*: A method where only robot and haptic data are used in our model.

- *Using Robot and Depth Data (R+D)*: A method where only robot and depth data are used in our model.

- *Using All Data (R+H+D)*: A method that fully uses all types of multimodal human response data.

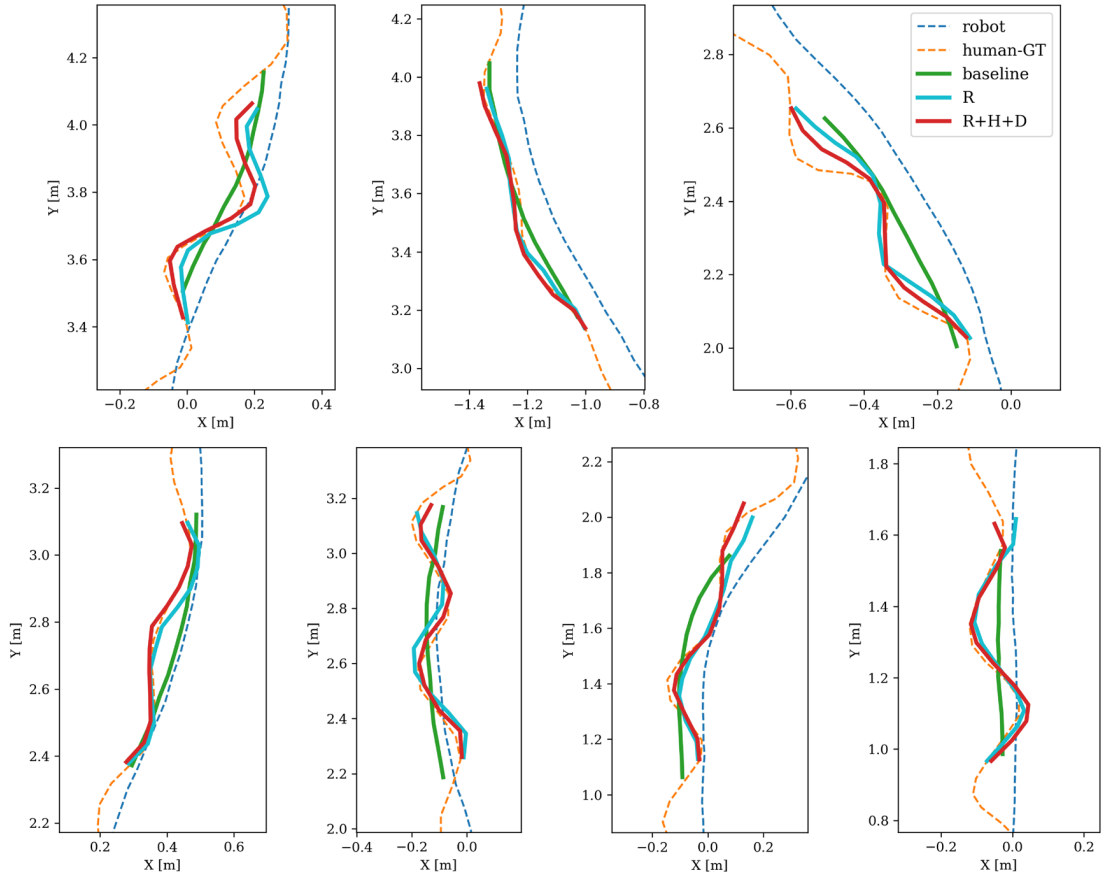

Figure 5. Examples of human path prediction results using our models. The dotted lines represent the entire trajectory of the actual robot and human movement. The bold lines represent the predicted trajectory of each method during the 3-second prediction time.

To generate the results of our model, a leave-one-out cross validation method is used; we repeat training 5 times with data from 4 participants and test on the remaining data set. Additional results are obtained to verify that this method can be used to predict the trajectories of other people whose data are not used for training. To do this, we repeat training 6 times with data from 5 participants and test on the other remaining participant's data.

## V. RESULTS

Table I contains the quantitative results of the methods we compared. The best results for each metric are in bold and underlined. For the baseline method, the result is calculated with the value $n = 8$, which means the person is assumed to move being attached to the position where was the robot 8 time-steps (i.e. 2 seconds) before. The results show that our proposed model greatly improves prediction performance compared to the baseline method, even when only using the robot data (R). When using the full multimodal data (R+H+D), which includes the haptic and depth data of the human, the results show even better prediction performance over the baseline with the best average values (i.e. MDE and MAE). Comparing the effectiveness of using the haptic (R+H) and depth data (R+D), the depth data model achieves slightly better performance, and both combination models show lower prediction error compared to using only the robot data (R). In particular, the usage of depth data shows the best performance in terms of final values (i.e. FDE and FAE).

Examples of human path prediction results using the baseline and our models (especially R and R+H+D) are presented in Fig. 5. This demonstrates the qualitative difference between the baseline method and our model results in terms of human body fluctuation in response to stepping on the person's left and right foot, which is also described in [26]. While the baseline method predicts the human's movement only as a trend line, our models provide a relatively accurate prediction of how the human body will move for 3 seconds based on the observation of human movement for the previous 2 seconds. This is valuable in that our models allow the robot to gather more information about the following pedestrian. For example, an assistive robot or a robot worker can identify in advance which stepping action the user will take at some point in the future. The prediction samples also indicate that the R+H+D method predicts a more accurate path compared to the R method, while both methods predict human body fluctuation.

TABLE I. PREDICTION ERRORS OF OUR METHODS

| Methods | Distance Error [m] | | Angle Error [deg] | |
| --- | --- | --- | --- | --- |
| | *MDE* | *FDE* | *MAE* | *FAE* |
| Baseline | 0.0813 | 0.0963 | 7.543 | 8.978 |
| R | 0.0434 | 0.0599 | 4.282 | 5.977 |
| R+H | 0.0423 | 0.0584 | 4.123 | 5.719 |
| R+D | 0.0420 | **<u>0.0574</u>** | 4.112 | **<u>5.685</u>** |
| R+H+D | **<u>0.0418</u>** | 0.0578 | **<u>4.099</u>** | 5.714 |

TABLE II. PREDICTION ERRORS OF OUR METHODS
(USING DATA OF THE PARTICIPANT NOT USED FOR TRAINING)

| Methods | Distance Error [m] | | Angle Error [deg] | |
| --- | --- | --- | --- | --- |
| | *MDE* | *FDE* | *MAE* | *FAE* |
| Baseline | 0.0813 | 0.0963 | 7.543 | 8.978 |
| R | 0.0498 | **0.0698** | 4.908 | **6.855** |
| R+H | **0.0493** | 0.0700 | 4.936 | 7.054 |
| R+D | 0.0508 | 0.0720 | 4.989 | 6.999 |
| R+H+D | 0.0506 | 0.0721 | **4.907** | 6.999 |

We further calculated the prediction error using data of the participant whose data was not used for training by changing the training method, as described in Section IV-B. Table II contains the quantitative results thereof. The baseline method result is the same as that of Table I, which is calculated assuming *n* = 8, as in the previous result. According to this, although prediction using our model does not reach the performance of Table I, it still shows better performance compared to the baseline method. This indicates that even with a completely new person's data, using our model allows a somewhat accurate path prediction. However, there is no clear performance improvement using multimodal data, and even the use of only robot data shows the best result in some metrics. This may be due to the lack of participant data or complex inter-participant dependency of the haptic and the depth of human response data.

## VI. CONCLUSION

In this paper, we present a method based on recurrent neural networks, which predicts the future movements of humans following a robotic guide without using sight. We use multimodal human response data obtained during physical interactions between the human and the robot. Our model outperforms the baseline result when using only the robot movement information, and it shows even better prediction performance when haptic and depth data are added. Our work can be helpful for robots that collaborate with humans to improve social intelligence. Specifically, in that it successfully predicts even the side-to-side fluctuation of the pedestrian, the robot could potentially understand a following human's walking traits. In addition, our model demonstrates that we can simply accept a new multimodal human response; therefore, the model can be further improved, e.g., having longer and more accurate prediction capability, by adding other response data.